\documentclass{article}

\usepackage[numbers]{natbib}

\usepackage[final]{neurips_2023_ml4ps}

\usepackage[utf8]{inputenc} 
\usepackage[T1]{fontenc}    
\usepackage{hyperref}       
\usepackage{url}            
\usepackage{booktabs}       
\usepackage{amsfonts}       
\usepackage{nicefrac}       
\usepackage{microtype}      
\usepackage{xcolor}         

\usepackage{amsmath,amssymb,amsfonts}
\usepackage{graphicx}
\usepackage{textcomp}

\usepackage{booktabs}

\usepackage{enumitem} 

\usepackage{subcaption} 
\usepackage{algorithm}
\usepackage{algpseudocode}
\algrenewcommand\algorithmicrequire{\textbf{Input:}}
\algrenewcommand\algorithmicensure{\textbf{Output:}}
\usepackage{tabularx}
\makeatletter
\newcommand{\multiline}[1]{%
  \begin{tabularx}{\dimexpr\linewidth-\ALG@thistlm}[t]{@{}X@{}}
    #1
  \end{tabularx}
}
\makeatother

\DeclareMathOperator*{\argmin}{arg\,min}

\def\Vec#1{{\boldsymbol{#1}}} 
\def\Mat#1{{\boldsymbol{#1}}} 

\usepackage{accents}
\newcommand{\leftup}[1]{\accentset{\leftharpoonup}{#1}}

\title{Application of Zone Method based Physics-Informed Neural Networks in Reheating Furnaces}   

\author{%
  Ujjal Kr Dutta  \\
  University College London\\
  WC1E 6BT, UK \\
  \And
  Aldo Lipani \\
  University College London \\
  WC1E 6BT, UK \\
  \And
  Chuan Wang \\
  Swerim AB \\
  Box 812, SE-97125, Sweden \\
  \And
  Yukun Hu \\
  University College London \\
  WC1E 6BT, UK \\
  \texttt{yukun.hu@ucl.ac.uk} \\
}

\begin{document}

\maketitle

\begin{abstract}
Foundation Industries (FIs) constitute glass, metals, cement, ceramics, bulk chemicals, paper, steel, etc. and provide crucial, foundational materials for a diverse set of economically relevant industries: automobiles, machinery, construction, household appliances, chemicals, etc. Reheating furnaces within the manufacturing chain of FIs are energy-intensive. Accurate and real-time prediction of underlying temperatures in reheating furnaces has the potential to reduce the overall heating time, thereby controlling the energy consumption for achieving the Net-Zero goals in FIs. In this paper, we cast this prediction as a regression task and explore neural networks due to their inherent capability of being effective and efficient, given adequate data. However, due to the infeasibility of achieving good-quality real data in scenarios like reheating furnaces, classical Hottel's zone method based computational model has been used to generate data for model training. To further enhance the Out-Of-Distribution generalization capability of the trained model, we propose a Physics-Informed Neural Network (PINN) by incorporating prior physical knowledge using a set of novel Energy-Balance regularizers.
\end{abstract}

\maketitle

\begin{figure*}[!tbh]
\centering
    \begin{subfigure}{.42\textwidth}
    	\centering
		\includegraphics[trim={0cm 0cm 0cm 0cm},clip,width=\textwidth]{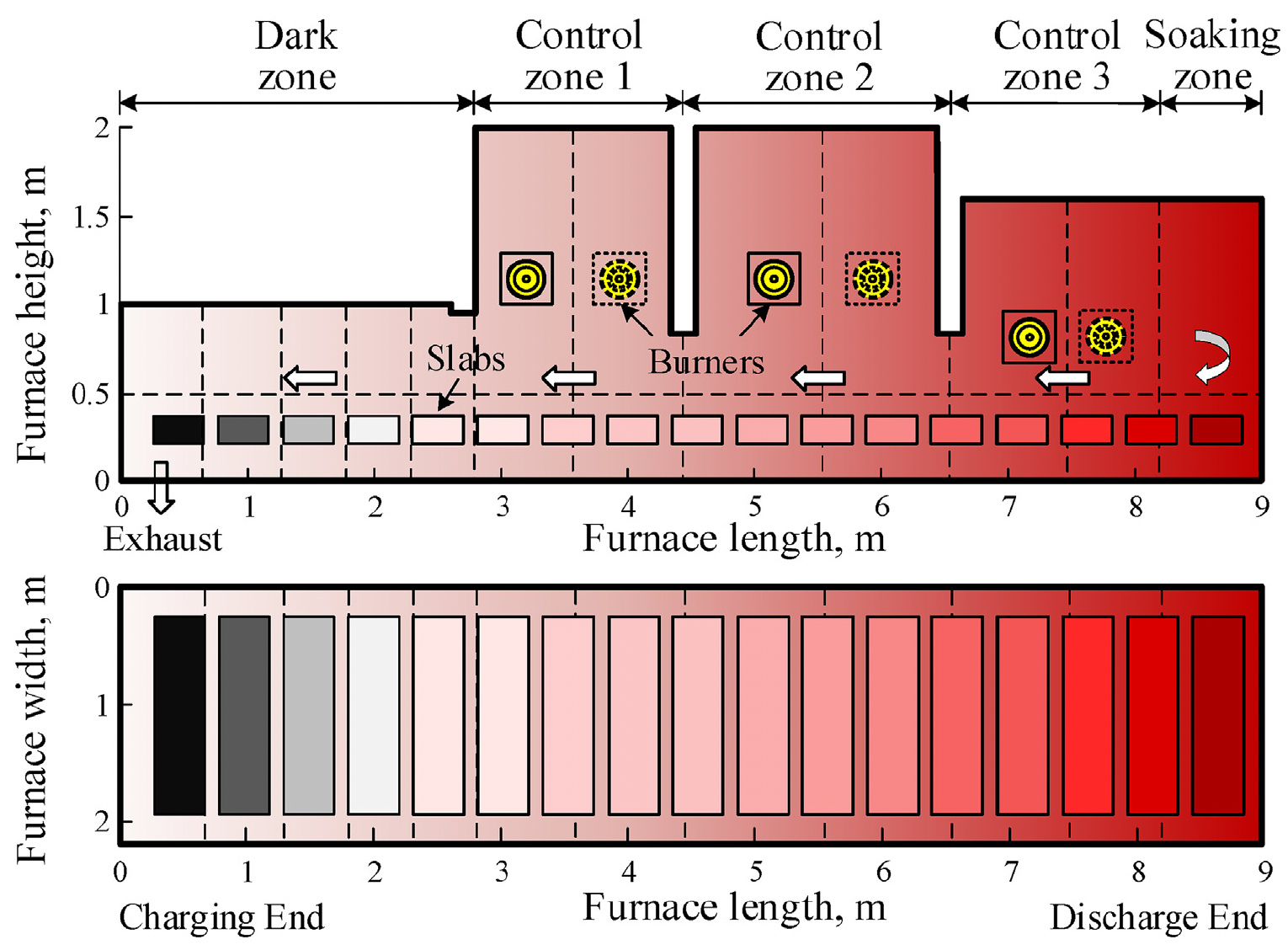}
		\caption{}
        \label{furnace_diag}
    \end{subfigure}
    \begin{subfigure}{.54\textwidth}
    	\centering
		\includegraphics[trim={0cm 0cm 0cm 0cm},clip,width=\textwidth]{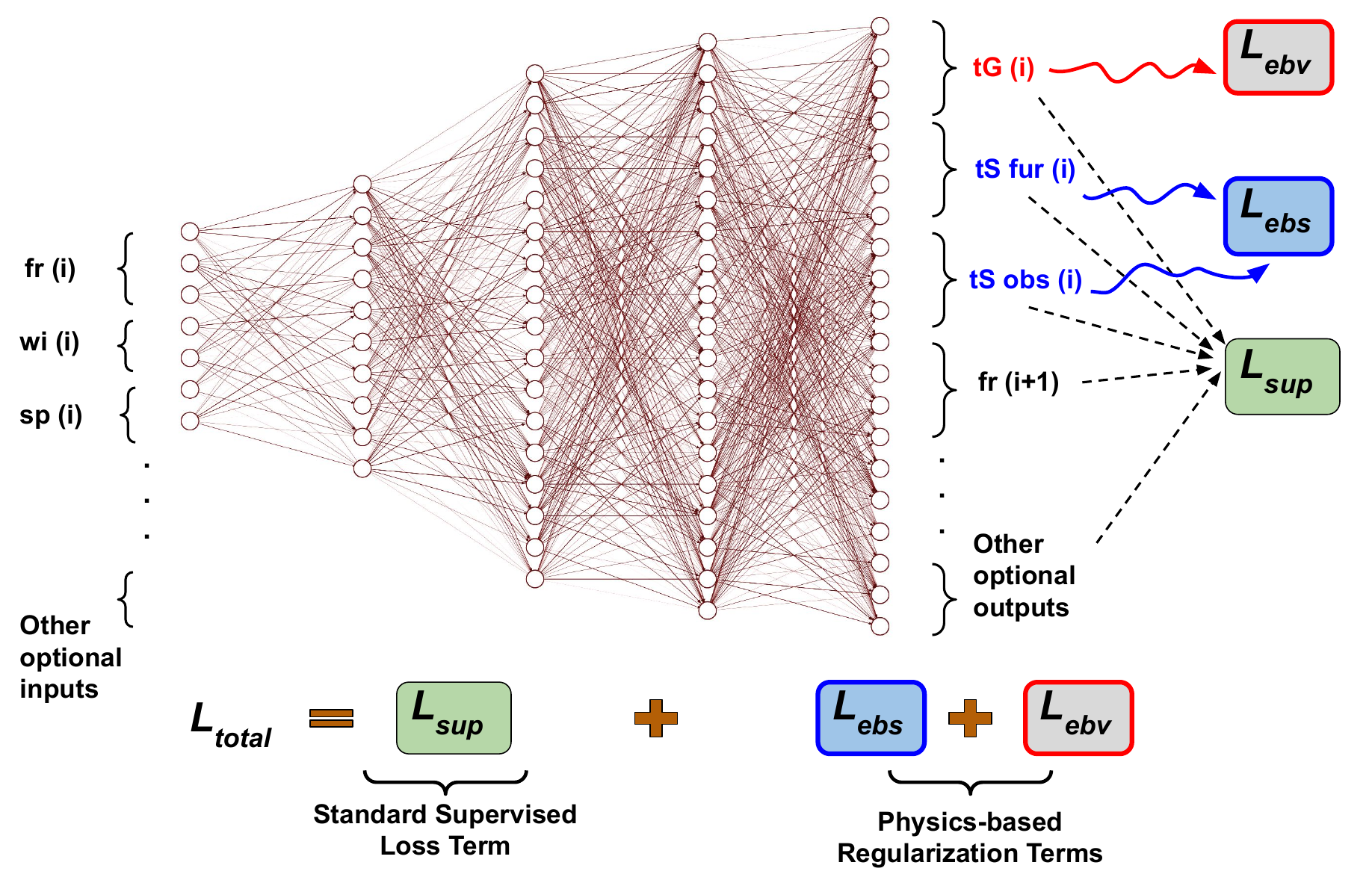}
		\caption{}
        \label{pi_mlp_illus}
    \end{subfigure}
    \caption{This figure is best viewed in color. Sub-figure (a) Illustration of a real-world furnace, and its subdivision as different zones. Image courtesy: \cite{hu2019modelling}. A darker shade of red indicates a higher temperature. Under normal conditions, temperature increases towards the discharge end. Sub-figure (b) Illustration of incorporating zone method based regularization to train a Physics-Informed Neural Network (PINN).}
    \label{abc}
\end{figure*}

\section{Proposed Method}

\textbf{Introduction and Motivation}: In this work, we tackle the key challenge of accurate and real-time temperature prediction in reheating furnaces, which are the energy-intensive bottlenecks common across the FIs. Available computational surrogate models based on Computational Fluid Dynamics (CFD) \cite{wehinger2019radiation, de2017methodology}, Discrete Element Method (DEM) \cite{emady2016prediction}, CFD-DEM hybrids \cite{oschmann2018novel}, Two Fluid Models (TFM) \cite{marti2015numerical}, etc. incur expensive and time-consuming data acquisition, design, optimization, and high inference times (of the order of tens of seconds, up to minutes). Deep Learning (DL) methods, on the other hand, owing to their accuracy and their inherently faster inference times (often only in the order of milliseconds), are suitable candidates for real-time prediction.

But unlike other industry settings, only a limited number of thermo-couples could be physically deployed within furnaces, making it infeasible to get even near-optimal, large-scale, good-quality real-world data to train a data-hungry DL model. The classical Hottel's zone method \cite{hottel1958radiant, hottel1967radiative, yuen1997zonal, hu2016development, hu2019modelling} provides an elegant way (and superior, as studied by Yuen and Takara \cite{yuen1997zonal}) to model the physical phenomenon in high-temperature processes inside reheating furnaces.

\textbf{Zone method for DL}: In a real-world furnace as in Figure \ref{furnace_diag}, the release of combustion materials (by burners, controlled via their firing rates) and movement of objects to be heated (slabs, or obstacles) from the left to the right (discharge end, with higher temperature), causes energy and mass flow. The zone method mathematically models this by dividing the furnace into a set of zones: i) G: Gas/ volume and ii) S: Surface (consisting of furnace walls \textit{fur} and obstacle surfaces \textit{obs}). The radiation interchange ($\leftharpoonup$ indicates the direction of flow) among all possible pairs $(i,j)$ of zones: Gas to Gas ($\leftup{\Mat{G_iG_j}}$), Surface to Surface ($\leftup{\Mat{S_iS_j}}$), Surface to Gas ($\leftup{\Mat{G_iS_j}}$), and Gas to Surface ($\leftup{\Mat{S_iG_j}}$), can be modeled along with a set of \textbf{Energy-Balance (EB) equations}.

Hu et al. \cite{hu2016development} has proposed a computational model of the zone method, which though highly accurate, is slower for real-time prediction. We use it, and simulate an offline, IID data set $\mathcal{X}_{IID}$=$\{ (\Vec{x}^{(i)}, \Vec{y}^{(i)} ) \}_{i=1}^N$ for DL training, by following their algorithmic flow. The advantage of this model is that the required input entities (e.g., ambient temperatures, set point temperatures, firing rates, walk-interval) are readily available without dependency on the physical placement of sensors in every relevant location where we want to collect data in the real-world.

We study the following two settings:
\setlist{nolistsep}
\begin{enumerate}[noitemsep]
    \item \textbf{Input Setting 1 - Without previous temperatures in the input vector:} Here, for time step instance $i$, we set: $\Vec{x}^{(i)}$ = $[fr(i)^\top, wi(i)^\top,sp(i)^\top]^\top$, and \\$\Vec{y}^{(i)}$ = $[tG(i)^\top, tS\textrm{ }fur(i)^\top, tS\textrm{ }obs(i)^\top, fr(i+1)^\top]^\top$.
    \item \textbf{Input Setting 2 - With previous temperatures in the input vector:} Here, for time step instance $i$, we set: $\Vec{x}^{(i)}$ = $[fr(i)^\top, wi(i)^\top,sp(i)^\top, tG(i-1)^\top,tS\textrm{ }fur(i-1)^\top,tS\textrm{ }obs(i-1)^\top]^\top$, and \\$\Vec{y}^{(i)}$ = $[tG(i)^\top, tS\textrm{ }fur(i)^\top, tS\textrm{ }obs(i)^\top, fr(i+1)^\top]^\top$.
\end{enumerate}
Here, $fr(i)$, $wi(i)$, $sp(i)$, $tG(i)$, $tS\textrm{ }fur(i)$ and $tS\textrm{ }obs(i)$ are respectively the vectors containing firing rates, walk-interval, set point temperatures, gas zone temperatures, surface zone temperatures for furnace walls, and surface zone temperatures for obstacles for a time step $i$. Also, $tG(i-1)$, $tS\textrm{ }fur(i-1)$, and $tS\textrm{ }obs(i-1)$ are respective vectors containing the corresponding temperatures from the previous time step. $fr(i+1)$ is a vector containing firing rates for the next time step.

Using $\mathcal{X}_{IID}$=$\{ (\Vec{x}^{(i)}, \Vec{y}^{(i)} ) \}_{i=1}^N$, we can estimate parameters $\theta$ of a Multi-Layer Perceptron (MLP) model $f_\theta(.)$ by training it to predict $\Vec{y}^{(i)}$ given $\Vec{x}^{(i)}$, for all $i$, as:
\begin{equation}
    \label{loss_mlp}
    \theta^* \leftarrow \argmin_{\theta} \mathbb{E}_{(\Vec{x}^{(i)},\Vec{y}^{(i)}) \in \mathcal{X}_{IID}}[|| \Vec{y}^{(i)} - f_{\theta}(\Vec{x}^{(i)}) ||_2^2]
\end{equation}
Then, we can obtain the required values of temperatures by extracting them from $f_{\theta^*}(\Vec{x}^{(i)})$.

\textbf{Zone method based PINN}: DL models are not naturally good at generalizing to Out-Of-Distribution (OOD) instances \cite{domainbed_iclr}. In our context, such OOD data could belong to furnace configurations (operating conditions) not seen during training. To tackle this, we propose employing a novel Physics-Informed Neural Network (PINN) model \cite{karniadakis2021physics} based on MLP. This is done by incorporating prior physical knowledge based on the zone method using a set of our novel proposed Energy-Balance regularizers.

To explain our PINN (see Figure \ref{pi_mlp_illus}), let us use eq(\ref{loss_mlp}) and denote: $\mathcal{L}_{sup} = \mathbb{E}_{(\Vec{x}^{(i)},\Vec{y}^{(i)}) \in \mathcal{X}_{IID}}[|| \Vec{y}^{(i)} - f_{\theta}(\Vec{x}^{(i)}) ||_2^2]$ as the standard \textit{supervised term}. Then, the overall PINN loss is formulated as:
\begin{equation}
    \label{ebv_ebs_mlp}
    \mathcal{L}_{total}=\mathcal{L}_{sup}+\lambda_{ebv}\mathcal{L}_{ebv}+\lambda_{ebs}\mathcal{L}_{ebs}
\end{equation}
Here, $\lambda_{ebv},\lambda_{ebs}>0$ are hyper-parameters corresponding to $\mathcal{L}_{ebv}$ and $\mathcal{L}_{ebs}$, such that $\mathcal{L}_{ebv}$=$|| \textrm{normalize}(\Vec{v}_g) ||_2^2$ is our proposed regularizer term corresponding to the \textbf{E}nergy-\textbf{B}alance equations for the \textbf{V}olume zones (\textbf{EBV}) using the zone method. Similarly, $\mathcal{L}_{ebs}$=$|| \textrm{normalize}(\Vec{v}_s) ||_2^2$ is our proposed regularizer term corresponding to the \textbf{E}nergy-\textbf{B}alance equations for the \textbf{S}urface zones (\textbf{EBS}). Normalizing an output vector of a neural network in a regression task is standard practice for ensuring convergence. In our work, we use: $\textrm{normalize}(\Vec{v}) = \Vec{v}/\textrm{max}(\Vec{v})$, where $\textrm{max}(\Vec{v})$ is the maximum value from among all components in $\Vec{v}$. We propose to represent $\Vec{v}_g$ and $\Vec{v}_s$ as:
\begin{equation}
    \label{eqn_vg_vs}
    \begin{split}
    & \Vec{v}_g = (\Vec{g}_{(g)arr} + \Vec{s}_{(g)arr} - 4\Vec{g}_{leave} + \Vec{h}_{g}) \in \mathbb{R}^{|G|}\\
    & \Vec{v}_s = (\Vec{s}_{(s)arr} + \Vec{g}_{(s)arr} - \Vec{s}_{leave} + \Vec{h}_{s}) \in \mathbb{R}^{|S|}
    \end{split}
\end{equation}
Here, $|G|/|S|$ denotes the number of Gas/ Surface zones. Intuitively, $\Vec{v}_g$ and $\Vec{v}_s$ are vector representatives corresponding to Energy-Balance equations for gas and surface zones respectively.

Having discussed $\Vec{v}_g$ and $\Vec{v}_s$, we now define the terms used to compute them. Let, $\Vec{g}_{(g)arr} \in \mathbb{R}^{|G|}$ be a vector whose $i^{th}$ entry represents the amount of radiation arriving at the $i^{th}$ gas zone from all the other gas zones, $\Vec{s}_{(g)arr} \in \mathbb{R}^{|G|}$, a vector whose $i^{th}$ entry represents the amount of radiation arriving at the $i^{th}$ gas zone from all the other surface zones, $\Vec{g}_{leave} \in \mathbb{R}^{|G|}$, a vector whose $i^{th}$ entry represents the amount of radiation leaving the $i^{th}$ gas zone, and $\Vec{h}_{g} \in \mathbb{R}^{|G|}$ a heat term. Also, let $T_{g,j}$ (or $T_{g}$) and $T_{s,j}$ (or $T_{s}$) denote the $j^{th}$ gas and surface zone temperatures respectively. Then, following EBV equations, the $i^{th}$ entries of $\Vec{g}_{(g)arr}$, $\Vec{s}_{(g)arr}$, $\Vec{g}_{leave}$ and $\Vec{h}_{g}$ can be computed as:
\begin{equation}
\label{ebv_eqns}
{
\footnotesize
\begin{aligned}
& \Vec{g}_{(g)arr}(i)=\sum_j^{|G|}\leftup{\Mat{G_iG_j}}\sigma T_{g,j}^4 \\
& \Vec{s}_{(g)arr}(i)=\sum_j^{|S|}\leftup{\Mat{G_iS_j}}\sigma T_{s,j}^4 \\
& \Vec{g}_{leave}(i)=\sum_n^{|N_g|} a_{g,n}(T_{g,i})k_{g,n} \sigma V_i T_{g,i}^4 \\
& \Vec{h}_{g}(i)=-(\dot{Q}_{conv})_i+(\dot{Q}_{fuel,net})_i+(\dot{Q}_{a})_i+\Vec{q}_i
\end{aligned}
}
\end{equation}
Here, the constants (known apriori) $(\dot{Q}_{conv})_i$, $(\dot{Q}_{fuel,net})_i$, and $(\dot{Q}_{a})_i$ respectively denote the convection heat transfer, heat release due to input fuel, and thermal input from air/ oxygen. An enthalpy vector $\Vec{q} \in \mathbb{R}^{|G|}$ is computed using the flow-pattern obtained via polynomial curve fitting during simulation. $\sigma$ is the Stefan-Boltzmann constant, $V_i$ is volume of $i^{th}$ gas zone.

Let, $\Vec{s}_{(s)arr} \in \mathbb{R}^{|S|}$, be a vector whose $i^{th}$ entry represents the amount of radiation arriving at the $i^{th}$ surface zone from all the other surface zones, $\Vec{g}_{(s)arr} \in \mathbb{R}^{|S|}$, a vector whose $i^{th}$ entry represents the amount of radiation arriving at the $i^{th}$ surface zone from all the other gas zones, $\Vec{s}_{leave} \in \mathbb{R}^{|S|}$, a vector whose $i^{th}$ entry represents the amount of radiation leaving the $i^{th}$ surface zone, and $\Vec{h}_{s} \in \mathbb{R}^{|S|}$ a heat term. Then, following EBS equations, the $i^{th}$ entries of $\Vec{s}_{(s)arr}$, $\Vec{g}_{(s)arr}$, $\Vec{s}_{leave}$ and $\Vec{h}_{s}$ can be computed as:
\begin{equation}
\label{ebs_eqns}
{
\footnotesize
\begin{aligned}
& \Vec{s}_{(s)arr}(i)=\sum_j^{|S|}\leftup{\Mat{S_iS_j}}\sigma T_{s,j}^4 \\
& \Vec{g}_{(s)arr}(i)=\sum_j^{|G|}\leftup{\Mat{S_iG_j}}\sigma T_{g,j}^4 \\
& \Vec{s}_{leave}(i)=A_i\epsilon_i\sigma T_{s,i}^4 \\
& \Vec{h}_{s}(i)=A_i(\dot{q}_{conv})_i-\dot{Q}_{s,i}
\end{aligned}
}
\end{equation}
For a surface zone $i$, the constants (known apriori) $A_i(\dot{q}_{conv})_i$ and $\dot{Q}_{s,i}$ respectively denote the heat flux to the surface by convection and heat transfer from it to the other surfaces. Here, $A_i$ is the area, and $\epsilon_i$ is the emissivity of the $i^{th}$ surface zone. In eq(\ref{ebv_eqns}), since the computations are being done for learning the gas zone related terms, the $T_g$ terms after being obtained from $f_{\theta}(\Vec{x})$ ($\Vec{x}$: input tensor to the PINN) are kept associated with the computational graph for back-propagating, but not the $T_s$ terms. The reverse is true in eq(\ref{ebs_eqns}) where we are learning for the surface zone related terms, i.e., $T_s$ terms are kept in the computational graph for back-propagating, but not $T_g$ terms. In addition, eq(\ref{ebv_eqns}) and eq(\ref{ebs_eqns}) also contain the DFAs ($\leftup{\Mat{GS}} \in \mathbb{R}^{|G|\times |S|}$, $\leftup{\Mat{SS}} \in \mathbb{R}^{|S|\times |S|}$, $\leftup{\Mat{GG}} \in \mathbb{R}^{|G|\times |G|}$, and $\leftup{\Mat{SG}} \in \mathbb{R}^{|S|\times |G|}$) and terms such as $a_{g,n}(T_{g,i}), k_{g,n}$, which can be referred from Hu et al. \cite{hu2016development}.

\section{Experimental Results}

\begin{table*}[!t]
\begin{minipage}[b]{0.53\textwidth}

\centering
\caption{Comparison of our proposed method against naive baseline and MLP without physics-based regularizer, in an IID evaluation setting.}
\label{iid_eval_naive_MLP}
\resizebox{\textwidth}{!}{%
\begin{tabular}{@{}cccc@{}}
\toprule
\multicolumn{4}{c}{\textbf{Without previous temperatures as inputs}} \\ \midrule
\multicolumn{1}{c|}{} &
  \multicolumn{2}{c|}{Baseline Methods} &
  \begin{tabular}[c]{@{}c@{}}Proposed \\ Physics-Informed Method\end{tabular} \\ \midrule
\multicolumn{1}{c|}{\begin{tabular}[c]{@{}c@{}}Performance\\ Metric\end{tabular}} &
  Naive Avg &
  \multicolumn{1}{c|}{MLP Baseline} &
  EBV+EBS \\ \midrule
\multicolumn{1}{c|}{RMSE tG ($\downarrow$)} &
  58.63 &
  \multicolumn{1}{c|}{10.27} &
  \textbf{10.04} \\
\multicolumn{1}{c|}{RMSE tS fur ($\downarrow$)} &
  53.03 &
  \multicolumn{1}{c|}{8.94} &
  \textbf{7.95} \\
\multicolumn{1}{c|}{RMSE tS obs  ($\downarrow$)} &
  68.19 &
  \multicolumn{1}{c|}{\textbf{30.94}} &
  31.64 \\
\multicolumn{1}{c|}{MAE tG ($\downarrow$)} &
  39.04 &
  \multicolumn{1}{c|}{7.31} &
  \textbf{7.19} \\
\multicolumn{1}{c|}{MAE tS fur ($\downarrow$)} &
  34.45 &
  \multicolumn{1}{c|}{5.97} &
  \textbf{5.58} \\
\multicolumn{1}{c|}{MAE tS obs ($\downarrow$)} &
  42.27 &
  \multicolumn{1}{c|}{\textbf{14.95}} &
  15.13 \\
\multicolumn{1}{c|}{$R^{2}$ tG ($\uparrow$)} &
  -0.031 &
  \multicolumn{1}{c|}{0.954} &
  \textbf{0.961} \\
\multicolumn{1}{c|}{$R^{2}$ tS fur ($\uparrow$)} &
  -0.042 &
  \multicolumn{1}{c|}{0.948} &
  \textbf{0.959} \\
\multicolumn{1}{c|}{$R^{2}$ tS obs ($\uparrow$)} &
  -0.065 &
  \multicolumn{1}{c|}{\textbf{0.886}} &
  0.885 \\
\multicolumn{1}{c|}{mMAPE fr ($\downarrow$)} &
  155.30 &
  \multicolumn{1}{c|}{7.41} &
  \textbf{6.84} \\ \midrule
\multicolumn{4}{c}{\textbf{With previous temperatures as inputs}} \\ \midrule
\multicolumn{1}{c|}{} &
  \multicolumn{2}{c|}{Baseline Methods} &
  \begin{tabular}[c]{@{}c@{}}Proposed \\ Physics-Informed Method\end{tabular} \\ \midrule
\multicolumn{1}{c|}{\begin{tabular}[c]{@{}c@{}}Performance\\ Metric\end{tabular}} &
  Naive Avg &
  \multicolumn{1}{c|}{MLP Baseline} &
  EBV+EBS \\ \midrule
\multicolumn{1}{c|}{RMSE tG ($\downarrow$)} &
  58.63 &
  \multicolumn{1}{c|}{5.75} &
  \textbf{4.91} \\
\multicolumn{1}{c|}{RMSE tS fur ($\downarrow$)} &
  53.03 &
  \multicolumn{1}{c|}{4.77} &
  \textbf{4.24} \\
\multicolumn{1}{c|}{RMSE tS obs  ($\downarrow$)} &
  68.19 &
  \multicolumn{1}{c|}{\textbf{17.18}} &
  17.39 \\
\multicolumn{1}{c|}{MAE tG ($\downarrow$)} &
  39.04 &
  \multicolumn{1}{c|}{3.21} &
  \textbf{3.01} \\
\multicolumn{1}{c|}{MAE tS fur ($\downarrow$)} &
  34.45 &
  \multicolumn{1}{c|}{3.09} &
  \textbf{2.74} \\
\multicolumn{1}{c|}{MAE tS obs ($\downarrow$)} &
  42.27 &
  \multicolumn{1}{c|}{\textbf{4.80}} &
  5.81 \\
\multicolumn{1}{c|}{$R^{2}$ tG ($\uparrow$)} &
  -0.031 &
  \multicolumn{1}{c|}{0.984} &
  \textbf{0.989} \\
\multicolumn{1}{c|}{$R^{2}$ tS fur ($\uparrow$)} &
  -0.042 &
  \multicolumn{1}{c|}{0.983} &
  \textbf{0.989} \\
\multicolumn{1}{c|}{$R^{2}$ tS obs ($\uparrow$)} &
  -0.065 &
  \multicolumn{1}{c|}{\textbf{0.966}} &
  \textbf{0.966} \\
\multicolumn{1}{c|}{mMAPE fr ($\downarrow$)} &
  155.30 &
  \multicolumn{1}{c|}{7.86} &
  \textbf{6.87} \\ \bottomrule
\end{tabular}%
}

\end{minipage}
\hspace{0.3cm}
\begin{minipage}[b]{0.45\textwidth}

\centering
\caption{Comparison of our proposed method against MLP without physics-based regularizer, in an auto-regressive evaluation setting.}
\label{ar_eval_MLP}
\resizebox{\textwidth}{!}{%
\begin{tabular}{@{}cccc@{}}
\toprule
\multicolumn{4}{c}{\textbf{Without previous temperatures as inputs}} \\ \midrule
\multicolumn{1}{c|}{Metric/ Method} &
  MLP &
  \multicolumn{1}{c|}{EBV+EBS} &
  \begin{tabular}[c]{@{}c@{}}EBV+EBS\\ improvement\\ over MLP (in \%)\end{tabular} \\ \midrule
\multicolumn{1}{c|}{RMSE tG ($\downarrow$)} &
  28.6 &
  \multicolumn{1}{c|}{\textbf{27.3}} &
  4.2 \\
\multicolumn{1}{c|}{RMSE tS fur ($\downarrow$)} &
  10.1 &
  \multicolumn{1}{c|}{\textbf{9.6}} &
  4.8 \\
\multicolumn{1}{c|}{RMSE tS obs  ($\downarrow$)} &
  \textbf{42.7} &
  \multicolumn{1}{c|}{44.0} &
  -3.1 \\
\multicolumn{1}{c|}{MAE tG ($\downarrow$)} &
  17.1 &
  \multicolumn{1}{c|}{\textbf{16.1}} &
  5.8 \\
\multicolumn{1}{c|}{MAE tS fur ($\downarrow$)} &
  7.8 &
  \multicolumn{1}{c|}{\textbf{7.3}} &
  6.5 \\
\multicolumn{1}{c|}{MAE tS obs ($\downarrow$)} &
  \textbf{20.0} &
  \multicolumn{1}{c|}{20.2} &
  -1.1 \\
\multicolumn{1}{c|}{mMAPE fr ($\downarrow$)} &
  69.2 &
  \multicolumn{1}{c|}{\textbf{63.5}} &
  8.2 \\ \midrule
\multicolumn{4}{c}{\textbf{With previous temperatures as inputs}} \\ \midrule
\multicolumn{1}{c|}{Metric/ Method} &
  MLP &
  \multicolumn{1}{c|}{EBV+EBS} &
  \begin{tabular}[c]{@{}c@{}}EBV+EBS\\ improvement\\ over MLP (in \%)\end{tabular} \\ \midrule
\multicolumn{1}{c|}{RMSE tG ($\downarrow$)} &
  74.1 &
  \multicolumn{1}{c|}{\textbf{36.8}} &
  50.3 \\
\multicolumn{1}{c|}{RMSE tS fur ($\downarrow$)} &
  74.5 &
  \multicolumn{1}{c|}{\textbf{25.8}} &
  65.4 \\
\multicolumn{1}{c|}{RMSE tS obs  ($\downarrow$)} &
  83.3 &
  \multicolumn{1}{c|}{\textbf{65.3}} &
  21.5 \\
\multicolumn{1}{c|}{MAE tG ($\downarrow$)} &
  48.8 &
  \multicolumn{1}{c|}{\textbf{29.3}} &
  39.9 \\
\multicolumn{1}{c|}{MAE tS fur ($\downarrow$)} &
  49.7 &
  \multicolumn{1}{c|}{\textbf{20.8}} &
  58.2 \\
\multicolumn{1}{c|}{MAE tS obs ($\downarrow$)} &
  53.6 &
  \multicolumn{1}{c|}{\textbf{42.0}} &
  21.6 \\
mMAPE fr ($\downarrow$) &
  96.2 &
  \textbf{40.6} &
  57.8 \\ \bottomrule
\end{tabular}%
}
\end{minipage}
\end{table*}

\textbf{PINN vs MLP vs Naive Baseline:} In our experiments, we compare our proposed PINN against a baseline MLP with the same architecture as our PINN, but without the physics-based EB regularizers (architecture and training details in appendix). We also compare a naive baseline, which, for a test instance, simply predicts the average value of a target variable using the training data. To evaluate the methods, we make use of an IID dataset (details in Appendix). For each test instance, we have input and output ground-truth values. We cast the prediction as a regression problem, and hence we can make use of the following standard regression performance evaluation metrics: Root Mean Squared Error (RMSE), Mean Absolute Error (MAE), and Coefficient of determination ($R^2$). We separately compute a model's performance for gas zone temperatures, furnace surface zone temperatures and obstacle surface zone temperatures, thus resulting in the metrics RMSE tG, RMSE tS fur, RMSE tS obs, MAE tG, MAE tS fur, MAE tS obs, $R^{2}$ tG, $R^{2}$ tS fur, and $R^{2}$ tS obs.

We train all models in the IID training split, tune hyper-parameters using the validation split, and report all performance metrics on the test split. We also predict the next firing rates, and because they are in practice within the normalized range $[0,1]$ \cite{hu2016development}, we make use of a modified Mean Absolute Percentage Error (mMAPE), by adding a small value $\epsilon=0.05$ to the denominator of the MAPE computation (to scale up the metric values). A lower value of RMSE, MAE, and mMAPE indicates a better performance (indicated by $\downarrow$), while a higher value of $R^2$ indicates a better performance (indicated by $\uparrow$). The best obtained metric by a method shall be shown in \textbf{bold} in the result tables.

From Table \ref{iid_eval_naive_MLP}, we noticed the superior performance of our proposed PINN over the MLP, as it better respects the underlying physics. When previous temperatures are provided in the inputs, due to additional signals, performance of both improves, but ours becomes better. In all cases, both the MLP and our PINN performs significantly better than the naive baseline, thereby, highlighting that learning based methods indeed help in this scenario. 

We additionally perform evaluation on the test data in an Auto-Regressive (AR) manner: Only for the test data set, rearrange all the test instances of a furnace configuration according to their time step values. Now, use the model checkpoint obtained in an IID manner on the training data, to infer on the test data set for a configuration, and compute performance metrics. The metrics across all configurations in the test split are then averaged.  

Specifically, in the inference time step $t$ of the AR evaluation, instead of providing the trained model the input values of firing rates, gas and surface zone temperatures obtained during simulation, we rather use the model predicted values obtained in the inference time step $t-1$. This is similar to a real-world operation, where a deployed model would be expected to continuously predict different values and use them as inputs for the next model predictions.

In the IID evaluation setting (Table \ref{iid_eval_naive_MLP}), at each inference step, the data is sampled IID, and the model inputs are those that are obtained via the simulation, which would be correct, as per the zone model. However, in the AR evaluation, over time, the model inputs being provided by its own predictions done earlier, are prone to cumulative error propagation. Thus, we can see that in Table \ref{ar_eval_MLP}, the values of performance metrics have degraded compared to the metrics obtained in the IID evaluation setting (Table \ref{iid_eval_naive_MLP}). Even then, our PINN outperforms the baseline MLP, on an average.

Also, when previous temperatures are provided as inputs, this makes the AR evaluation more challenging. This is because there are now more input entities which could be predicted by the model sub-optimally. In this case, performance of the MLP deteriorates significantly. This might be because it merely learns to memorize the training data, without really understanding the underlying physical phenomenon. On the other hand, our method, being aware of the underlying physics, is more generalizable and hence performs significantly better than the MLP baseline (up to 50-65\% improvements). In the appendix, we provide additional experimental results, including the in-depth analysis of our PINN.

\textbf{Limitations of our PINN:} The original data obtained by the computational model of \cite{hu2016development} is time dependent in nature, i.e., data of a time step is dependent on the previous time step outputs. However, conforming to IID nature of data upon which standard ML/DL/MLP models are trained for regression, we simplified the structure of the data, to make it IID. However, one could keep the time dependent nature of the data intact and make use of a Recurrent Neural Network (RNN) to model the data during training. However, doing so is best justified as a separate future work.

At the same time, in Tables \ref{iid_eval_naive_MLP}-\ref{ar_eval_MLP}, we noticed that our PINN is unable to consistently improve the performance on the predicted obstacle surface zone temperatures ($tS\ obs$ performance). This is because our regularizer does not treat furnace and obstacle surfaces separately. Instead, the $\mathcal{L}_{ebs}$ term optimizes a collective loss across all surfaces (furnace and obstacles). The number of obstacle surfaces are higher in practice, than the number of furnace surfaces. Temperatures poorly predicted for a few of obstacle surfaces can increase the average errors. Our PINN regularizer does not proactively address this. Also, geometry and other features of the slabs (e.g., material quality) are not taken into account. While being geometry-agnostic is favorable to our framework in terms of simplicity and generalizability, there is a trade-off between formulating a generic vs specific model. These avenues can be addressed as independent works in the future.

\section{Conclusion}
While some loosely related prior works have touched upon aspects of radiative heat transfer, exchange area calculation \cite{ebrahimi2013zonal, melot2011comparison}, genetic algorithm for nonlinear dynamic systems \cite{eng_genes}, neural network for absorption coefficients \cite{rad_nnet_2009}, view factor modeling with DEM-based simulations \cite{tausendschon2021deep}, near-field heat transfer or close regime \cite{garcia2021deep}, and some on non neural network based temperature profiling in reheating furnaces \cite{kim2000prediction, jang2010investigation, kim2007heat, tang2017modeling, hu2018nonlinear, li2023novel, zanoli2023multi}, casting the temperature prediction task in reheating furnaces as a regression task, and modeling via explicit physics-constrained regularizers as done in our work, is a first of its kind. In the future, our work could be extended for newer avenues, such as incorporating additional furnace geometries via transfer learning and continual learning.

\section*{Acknowledgments}
The authors wish to acknowledge the Transforming Foundation Industries Network+ (EPSRC grant EP/V026402/1) for funding this work. Corresponding author's email: yukun.hu@ucl.ac.uk.

{\footnotesize
\bibliographystyle{unsrt}
\bibliography{EBVS_NeurIPS23}
}

\section*{Appendix}

\begin{table*}[!tbh]
\centering
\caption{Dataset details. A total of 50 configurations have been used which are categorized as normal or abnormal.}
\resizebox{\textwidth}{!}{%
\begin{tabular}{@{}ccccc@{}}
\toprule
\multicolumn{5}{c}{\textbf{Normal Behaviour Configurations (SP1\textless{}SP2\textless{}SP3)}}                                                                                                                                                                                                                                                                                                                                                                                                                                                                                                                                                                                                                                                                                                                                                                                                                                                                                                                                                                                                                                                                                                                                                                                                                                                                                                                    \\ \midrule
\multicolumn{1}{c|}{\textbf{Type 1 (Varying SP1 only)}}                                                                                                                                                                                                                                                                                                          & \multicolumn{1}{c|}{\textbf{Type 2 (Varying SP2 only)}}                                                                                                                                                                                                                                              & \multicolumn{1}{c|}{\textbf{Type 3 (Varying SP3 only)}}                                                                                                                                                                                                                                                                         & \multicolumn{2}{c}{\textbf{Type 4 (Varying WI only)}}                                                                                                                                                                                                                                                                                                                                                 \\ \midrule
\multicolumn{1}{c|}{\begin{tabular}[c]{@{}c@{}}905\_1220\_1250\_750.csv (Training)\\ 915\_1220\_1250\_750.csv (Val)\\ 925\_1220\_1250\_750.csv\\ 935\_1220\_1250\_750.csv (Training)\\ 945\_1220\_1250\_750.csv (Val)\\ 965\_1220\_1250\_750.csv\\ 975\_1220\_1250\_750.csv (Training)\\ 985\_1220\_1250\_750.csv (Val)\\ 995\_1220\_1250\_750.csv\end{tabular}} & \multicolumn{1}{c|}{\begin{tabular}[c]{@{}c@{}}955\_1170\_1250\_750.csv (Training)\\ 955\_1180\_1250\_750.csv (Val)\\ 955\_1190\_1250\_750.csv\\ 955\_1200\_1250\_750.csv (Training)\\ 955\_1210\_1250\_750.csv (Val)\\ 955\_1230\_1250\_750.csv\\ 955\_1240\_1250\_750.csv (Training)\end{tabular}} & \multicolumn{1}{c|}{\begin{tabular}[c]{@{}c@{}}955\_1220\_1230\_750.csv (Training)\\ 955\_1220\_1240\_750.csv (Val)\\ 955\_1220\_1250\_750.csv\\ 955\_1220\_1260\_750.csv (Training)\\ 955\_1220\_1270\_750.csv (Val)\\ 955\_1220\_1280\_750.csv\\ 955\_1220\_1290\_750.csv (Training)\\ 955\_1220\_1300\_750.csv\end{tabular}} & \multicolumn{2}{c}{\begin{tabular}[c]{@{}c@{}}955\_1220\_1250\_675.csv (Training)\\ 955\_1220\_1250\_690.csv (Val)\\ 955\_1220\_1250\_705.csv\\ 955\_1220\_1250\_720.csv (Training)\\ 955\_1220\_1250\_735.csv (Val)\\ 955\_1220\_1250\_765.csv\\ 955\_1220\_1250\_780.csv (Training)\\ 955\_1220\_1250\_795.csv (Val)\\ 955\_1220\_1250\_810.csv\\ 955\_1220\_1250\_825.csv (Training)\end{tabular}} \\ \midrule
\multicolumn{5}{c}{\textbf{Abnormal Behaviour Configurations/ Arbitrary SPs}}                                                                                                                                                                                                                                                                                                                                                                                                                                                                                                                                                                                                                                                                                                                                                                                                                                                                                                                                                                                                                                                                                                                                                                                                                                                                                                                                     \\ \midrule
\multicolumn{1}{c|}{\textbf{Type 1 (start@955-incr-dec/const)}}                                                                                                                                                                                                                                                                                                  & \multicolumn{1}{c|}{\textbf{Type 2 (start@1220-incr-dec)}}                                                                                                                                                                                                                                           & \multicolumn{1}{c|}{\textbf{Type 3 (start@1220-dec-inc)}}                                                                                                                                                                                                                                                                       & \multicolumn{1}{c|}{\textbf{Type 4 (start@1250-dec-inc)}}                                                                                                                                                    & \textbf{Type 5 (start@1250-dec-inc)}                                                                                                                                                   \\ \midrule
\multicolumn{1}{c|}{\begin{tabular}[c]{@{}c@{}}955\_1220\_1200\_750.csv (Training)\\ 955\_1220\_1210\_750.csv (Val)\\ 955\_1220\_1220\_750.csv\\ 955\_1250\_1220\_750.csv (Training)\\ 955\_1250\_1220\_765.csv (Val)\\ 955\_1250\_1250\_750.csv\\ 955\_1260\_1250\_750.csv (Training)\\ 955\_1270\_1250\_750.csv\end{tabular}}                                  & \multicolumn{1}{c|}{\begin{tabular}[c]{@{}c@{}}1220\_1250\_955\_750.csv (Training)\\ 1220\_1250\_955\_795.csv\end{tabular}}                                                                                                                                                                          & \multicolumn{1}{c|}{\begin{tabular}[c]{@{}c@{}}1220\_955\_1250\_750.csv (Training)\\ 1220\_955\_1250\_780.csv\end{tabular}}                                                                                                                                                                                                     & \multicolumn{1}{c|}{\begin{tabular}[c]{@{}c@{}}1250\_955\_1220\_750.csv (Training)\\ 1250\_955\_1220\_825.csv\end{tabular}}                                                                                  & \begin{tabular}[c]{@{}c@{}}1250\_1220\_955\_750.csv (Training)\\ 1250\_1220\_955\_810.csv\end{tabular}                                                                                 \\ \bottomrule
\end{tabular}%
}
\label{dataset_details}
\end{table*}

\textbf{Data set:} For the IID data set generation, we make use of a FORTRAN code provided by the authors of \cite{hu2019modelling}, to represent various furnace configurations of the real-world furnace shown in Figure \ref{furnace_diag}. We consider 50 different configurations, and create disjoint train-val-test splits in such a way that there is no overlap in the data across different splits. Also, each configuration could belong to either of train/val/test split. As val/test data belong to furnace configurations different from that of training, it naturally makes the test data OOD in nature. Each configuration can be defined by set point temperatures and the walk-interval. Set point temperatures are essentially the desired temperatures that the furnace is expected to achieve at different stages/ zones.

We represent a configuration as: \texttt{SP1\_SP2\_SP3\_WI}, where \texttt{SP1}, \texttt{SP2}, \texttt{SP3} and \texttt{WI} respectively denote the set point 1, set point 2, set point 3, and walk interval. Note that we consider configurations with both normal conditions (\texttt{SP1<SP2<SP3}, as naturally occurring in practice), as well as abnormal ones (arbitrary set points). The details are present in Table \ref{dataset_details}. Here, each configuration is represented by a .csv file containing 1500 time steps (and with the appropriate training/val label in parenthesis, and no label for a test split). Within a configuration, each time step is sampled with a 15s delay, to account for conduction analysis.

In Algorithm \ref{data_gen_algo}, we outline the key steps required in the data generation step, for a particular configuration. Please refer \cite{hu2016development} for details on the flow. Here, the major entities as discussed in our formulation are mentioned: $fr(t), wi(t) ,sp(t), tG(t),tS\textrm{ }fur(t),tS\textrm{ }obs(t)$. In addition, we also make use of auxiliary entities such as enthalpy $\Vec{q}^{(t)}$, heat-flux $\Vec{w}^{(t)}$, and node temperatures $\Vec{n}^{(t)}$. 

\begin{algorithm}[H]
\begin{scriptsize}
\caption{Data generation algorithm}
\label{data_gen_algo}
\begin{algorithmic}[1]
\State Initialize a furnace configuration via set points and walk interval.
\State Initialize $\mathcal{X}=\{  \}$, $T>0$ (max no. of steps).
\State Initialize $tG(0), tS\textrm{ }fur(0),tS\textrm{ }obs(0)$ with ambient temperatures, and $fr(0)$.
\For{t=1 {\bfseries to} $T$} \Comment{t: time step}
\State $fr(t) \leftarrow \textrm{update firing rates}(fr(t-1), \textrm{set point temperatures}, tG(t-1), tS\textrm{ }fur(t-1),tS\textrm{ }obs(t-1) )$
\State $\Vec{q}^{(t)} \leftarrow \textrm{Enthalpy}(\textrm{Flow-pattern}(fr(t)))$
\State $\leftup{\Mat{GG}}^{(t)}, \leftup{\Mat{GS}}^{(t)}, \leftup{\Mat{SG}}^{(t)}, \leftup{\Mat{SS}}^{(t)} \leftarrow \textrm{DFA}(tG(t-1), tS\textrm{ }fur(t-1),tS\textrm{ }obs(t-1), \Mat{GG}, \Mat{GS}, \Mat{SG}, \Mat{SS})$ 
\State $tG(t) \gets  \textrm{EBV} (\Vec{q}^{(t)}, \leftup{\Mat{GG}}^{(t)}, \leftup{\Mat{GS}}^{(t)}) $
\State $\Vec{w}^{(t)} \gets  \textrm{heat-transfer} (tG(t), tS\textrm{ }fur(t-1),tS\textrm{ }obs(t-1), \leftup{\Mat{SS}}^{(t)}, \leftup{\Mat{SG}}^{(t)}) $
\State $ tS\textrm{ }fur(t),tS\textrm{ }obs(t) \gets \textrm{EBS}(\textrm{conduction} (\Vec{w}^{(t)}) ) $, $ \Vec{n}^{(t)} \gets \textrm{conduction} (\Vec{w}^{(t)}) $
\State  $\mathcal{X}_t \leftarrow \{ fr(t), \textrm{Flow-pattern}(fr(t)), \Vec{q}^{(t)}, tG(t), tS\textrm{ }fur(t),tS\textrm{ }obs(t) , \Vec{w}^{(t)} , \Vec{n}^{(t)} \}$
\State $\mathcal{X} \leftarrow \mathcal{X} \cup \mathcal{X}_t$
\EndFor
\State \textbf{return} $\mathcal{X}$
\end{algorithmic}
\end{scriptsize}
\end{algorithm}

As the temperatures predicted in a time step influence the firing rates for the next time step, there is a time dependency among the data in $\mathcal{X}$. However, most standard off-the-shelf Machine Learning (ML)/ DL models suitable for regression require the data in an Independent and Identically Distributed (IID) format, that could be loaded in a tabular form (with each row being an instance and the columns representing the attributes). Thus, to convert $\mathcal{X}$ to $\mathcal{X}_{IID}$, we essentially add new columns and shift the entries.

Particularly, we add a new column \texttt{firing\_rates\_next} by shifting the original firing rates column a step back and then dropping the last row. Likewise, we add new columns for \textit{prev} temperatures by shifting the original temperature columns a step forward and then dropping the first row. After rearranging the data as IID, we consolidate all the 20 training, 12 validation, and 18 test configurations (with 1500 minus 2 time steps per configuration), resulting in 29960 train, 17976 val, and 26964 test time steps/ IID samples. The 2 time steps are subtracted to account for the shift operations discussed during the IID data creation.

\textbf{Training details and model architecture: }
We train our PINN model EBV+EBS for 10 epochs using PyTorch, with early stopping to avoid over-fitting. For the EB equations, we perform the same normalization for enthalpy, flux, and temperatures, as in the final neural network output as discussed earlier. We found a learning rate of 0.001 with Adam optimizer and batch size of 64 to be optimal, along with ReLU non-linearity.

We pick the [50,100,200] configuration for hidden layers, i.e., 3 hidden layers, with 50, 100, and 200 neurons respectively. We use $\lambda_{ebv}=\lambda_{ebs}=0.1$. In general, a value lesser than 1 is observed to be better, otherwise, the model focuses less on the regression task. Following are values of other variables: $|G|=24$, $|S|=178$ (76 furnace surface zones and 102 obstacle surface zones), $N_g=6$, and Stefan-Boltzmann constant=5.6687e-08. Unless otherwise stated, this is the setting we use to report any results for our method, for example, while comparing with other methods.

\begin{figure*}[!tbh]
\centering
	\begin{subfigure}{.32\textwidth}
    	\centering
		\includegraphics[trim={0cm 0cm 0cm 0cm},clip,width=\textwidth]{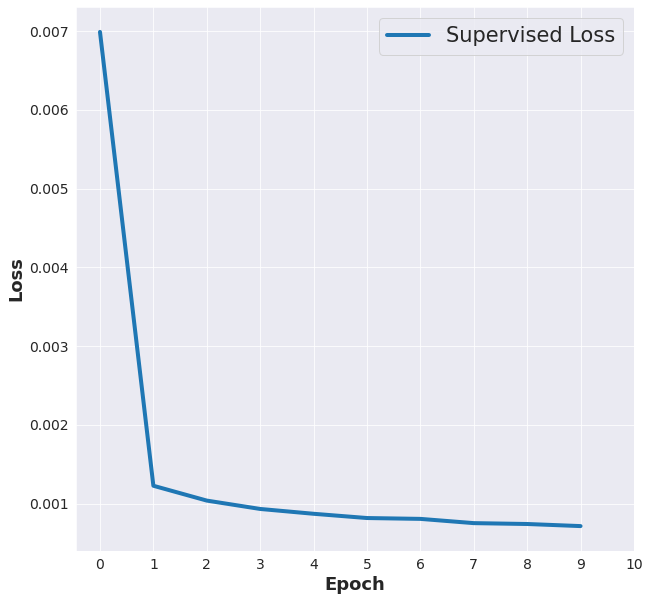}
		\caption{}
        \label{subfig_a}
    \end{subfigure}
	\begin{subfigure}{.32\textwidth}
    	\centering
		\includegraphics[trim={0cm 0cm 0cm 0cm},clip,width=\textwidth]{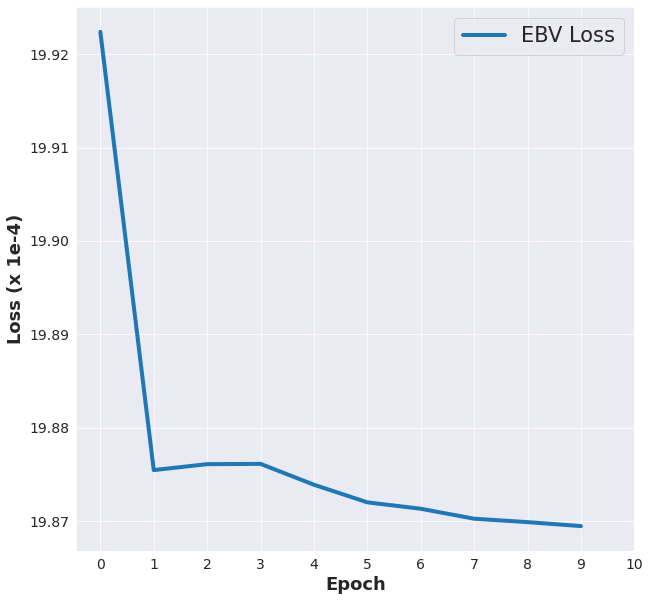}
		\caption{}
        \label{subfig_b}
    \end{subfigure}
    \begin{subfigure}{.32\textwidth}
    	\centering
		\includegraphics[trim={0cm 0cm 0cm 0cm},clip,width=\textwidth]{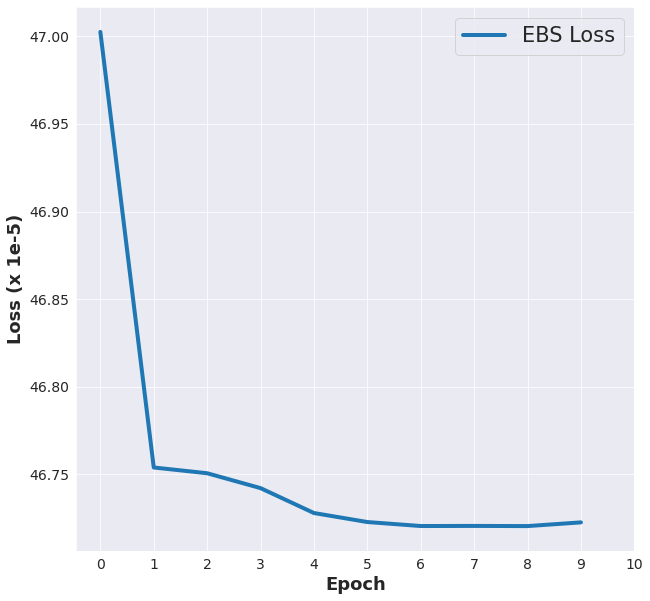}
		\caption{}
        \label{subfig_c}
    \end{subfigure}
    \caption{ Convergence behaviour of our method, considering: a) Supervised, b) EBV, and c) EBS terms.}
    \label{conv_ebvs}
\end{figure*}

\begin{table*}[!th]
\begin{minipage}[b]{0.55\textwidth}
\centering
\caption{Performance of EBV+EBS (ReLU) variant of our \\method against varying hidden layer configurations.}
\label{ablation_hidden}
\resizebox{\textwidth}{!}{%
\begin{tabular}{@{}c|ccccc@{}}
\toprule
\begin{tabular}[c]{@{}c@{}}Metric/ \\ Hidden layer \\ configuration\end{tabular} &
  {[}100{]} &
  {[}50,100{]} &
  \textbf{\begin{tabular}[c]{@{}c@{}}{[}50,100, \\ 200{]}\end{tabular}} &
  \begin{tabular}[c]{@{}c@{}}{[}50,100,\\ 200,200{]}\end{tabular} &
  \begin{tabular}[c]{@{}c@{}}{[}50,100,\\ 200,200,\\ 205,205{]}\end{tabular} \\ \midrule
RMSE tG ($\downarrow$)      & 11.64 & 17.25 & \textbf{10.04} & 10.84         & 14.27 \\
RMSE tS fur ($\downarrow$)  & 10.05 & 15.23 & 7.95           & \textbf{7.83} & 12.46 \\
RMSE tS obs  ($\downarrow$) & 34.82 & 37.62 & \textbf{31.64} & 33.57         & 36.42 \\
mMAPE fr ($\downarrow$)     & 8.76  & 9.15  & \textbf{6.84}  & 8.06          & 7.51  \\ \bottomrule
\end{tabular}%
}
\end{minipage}
\hspace{0.1cm}
\begin{minipage}[b]{0.43\textwidth}
\centering
\caption{Performance of the proposed EBV+EBS \\variant using different batch sizes.}
\label{ablation_batchsz}
\resizebox{\textwidth}{!}{%
\begin{tabular}{@{}c|ccc@{}}
\toprule
Metric &
  \begin{tabular}[c]{@{}c@{}}EBV+EBS\\ ReLU\\ bsz=32\end{tabular} &
  \begin{tabular}[c]{@{}c@{}}EBV+EBS\\ ReLU\\ bsz=64\end{tabular} &
  \begin{tabular}[c]{@{}c@{}}EBV+EBS\\ ReLU\\ bsz=128\end{tabular} \\ \midrule
RMSE tG ($\downarrow$)      & 12.70         & \textbf{10.04} & 10.73 \\
RMSE tS fur ($\downarrow$)  & 9.14          & \textbf{7.95}  & 9.69  \\
RMSE tS obs  ($\downarrow$) & 39.75         & \textbf{31.64} & 31.79 \\
mMAPE fr ($\downarrow$)     & \textbf{5.24} & 6.84           & 8.29  \\ \bottomrule
\end{tabular}%
}
\end{minipage}
\end{table*}

\begin{table*}[!th]
\begin{minipage}[b]{0.4\textwidth}
\centering
\caption{Effect of individual regularizer terms in our method.}
\label{ablation_ebv_ebs}
\resizebox{\textwidth}{!}{%
\begin{tabular}{@{}c|ccc@{}}
\toprule
Metric                      & EBV only      & EBS only & EBV+EBS        \\ \midrule
RMSE tG ($\downarrow$)      & 11.85         & 11.66    & \textbf{10.04} \\
RMSE tS fur ($\downarrow$)  & 10.36         & 11.07    & \textbf{7.95}  \\
RMSE tS obs  ($\downarrow$) & 32.46         & 32.04    & \textbf{31.64} \\
mMAPE fr ($\downarrow$)     & \textbf{6.42} & 7.53     & 6.84           \\ \bottomrule
\end{tabular}%
}
\end{minipage}
\hspace{0.1cm}
\begin{minipage}[b]{0.6\textwidth}
\centering
\caption{Performance of our method using different activation functions in the underlying network.}
\label{ablation_activationfn}
\resizebox{\textwidth}{!}{%
\begin{tabular}{@{}c|ccccc@{}}
\toprule
Metric &
  \begin{tabular}[c]{@{}c@{}}EBV+EBS\\ ReLU\end{tabular} &
  \begin{tabular}[c]{@{}c@{}}EBV+EBS\\ GeLU\end{tabular} &
  \begin{tabular}[c]{@{}c@{}}EBV+EBS\\ SiLU\end{tabular} &
  \begin{tabular}[c]{@{}c@{}}EBV+EBS\\ Hardswish\end{tabular} &
  \begin{tabular}[c]{@{}c@{}}EBV+EBS\\ Mish\end{tabular} \\ \midrule
RMSE tG ($\downarrow$)      & \textbf{10.04} & 13.57         & 10.07 & 15.26 & 10.16          \\
RMSE tS fur ($\downarrow$)  & 7.95           & 8.86          & 8.02  & 14.02 & \textbf{7.71}  \\
RMSE tS obs  ($\downarrow$) & 31.64          & 39.65         & 31.64 & 36.23 & \textbf{31.63} \\
mMAPE fr ($\downarrow$)     & 6.84           & \textbf{5.88} & 6.23  & 7.03  & 6.33           \\ \bottomrule
\end{tabular}%
}
\end{minipage}
\end{table*}

\begin{table*}[!tbh]
\centering
\caption{Comparison of our proposed method against classical ML baselines, in an IID evaluation setting.}
\label{iid_classical_ML}
\resizebox{0.6\textwidth}{!}{%
\begin{tabular}{@{}ccccc@{}}
\toprule
\multicolumn{5}{c}{\textbf{Without previous temperatures as inputs}} \\ \midrule
\multicolumn{1}{c|}{} &
  \multicolumn{3}{c|}{Classical ML Baselines} &
  \begin{tabular}[c]{@{}c@{}}Proposed \\ Physics-Informed Method\end{tabular} \\ \midrule
\multicolumn{1}{c|}{\begin{tabular}[c]{@{}c@{}}Performance\\ Metric\end{tabular}} &
  DT &
  RF &
  \multicolumn{1}{c|}{H-GBoost} &
  EBV+EBS \\ \midrule
\multicolumn{1}{c|}{RMSE tG ($\downarrow$)} &
  12.84 &
  12.24 &
  \multicolumn{1}{c|}{14.06} &
  \textbf{10.04} \\
\multicolumn{1}{c|}{RMSE tS fur ($\downarrow$)} &
  9.42 &
  8.97 &
  \multicolumn{1}{c|}{10.09} &
  \textbf{7.95} \\
\multicolumn{1}{c|}{RMSE tS obs  ($\downarrow$)} &
  42.86 &
  42.06 &
  \multicolumn{1}{c|}{42.73} &
  \textbf{31.64} \\
\multicolumn{1}{c|}{$R^{2}$ tG ($\uparrow$)} &
  0.943 &
  0.948 &
  \multicolumn{1}{c|}{0.925} &
  \textbf{0.961} \\
\multicolumn{1}{c|}{$R^{2}$ tS fur ($\uparrow$)} &
  0.951 &
  0.957 &
  \multicolumn{1}{c|}{0.934} &
  \textbf{0.959} \\
\multicolumn{1}{c|}{$R^{2}$ tS obs ($\uparrow$)} &
  0.788 &
  0.798 &
  \multicolumn{1}{c|}{0.763} &
  \textbf{0.885} \\
\multicolumn{1}{c|}{mMAPE fr ($\downarrow$)} &
  5.50 &
  5.30 &
  \multicolumn{1}{c|}{\textbf{2.32}} &
  6.84 \\ \midrule
\multicolumn{5}{c}{\textbf{With previous temperatures as inputs}} \\ \midrule
\multicolumn{1}{c|}{} &
  \multicolumn{3}{c|}{Classical ML Baselines} &
  \begin{tabular}[c]{@{}c@{}}Proposed \\ Physics-Informed Method\end{tabular} \\ \midrule
\multicolumn{1}{c|}{\begin{tabular}[c]{@{}c@{}}Performance\\ Metric\end{tabular}} &
  DT &
  RF &
  \multicolumn{1}{c|}{H-GBoost} &
  EBV+EBS \\ \midrule
\multicolumn{1}{c|}{RMSE tG ($\downarrow$)} &
  11.17 &
  6.96 &
  \multicolumn{1}{c|}{5.00} &
  \textbf{4.91} \\
\multicolumn{1}{c|}{RMSE tS fur ($\downarrow$)} &
  10.24 &
  6.15 &
  \multicolumn{1}{c|}{6.12} &
  \textbf{4.24} \\
\multicolumn{1}{c|}{RMSE tS obs  ($\downarrow$)} &
  43.05 &
  32.81 &
  \multicolumn{1}{c|}{23.01} &
  \textbf{17.39} \\
\multicolumn{1}{c|}{$R^{2}$ tG ($\uparrow$)} &
  0.925 &
  0.979 &
  \multicolumn{1}{c|}{\textbf{0.989}} &
  \textbf{0.989} \\
\multicolumn{1}{c|}{$R^{2}$ tS fur ($\uparrow$)} &
  0.915 &
  0.977 &
  \multicolumn{1}{c|}{0.983} &
  \textbf{0.989} \\
\multicolumn{1}{c|}{$R^{2}$ tS obs ($\uparrow$)} &
  0.729 &
  0.890 &
  \multicolumn{1}{c|}{0.937} &
  \textbf{0.966} \\
\multicolumn{1}{c|}{mMAPE fr ($\downarrow$)} &
  6.98 &
  8.09 &
  \multicolumn{1}{c|}{\textbf{0.76}} &
  6.87 \\ \bottomrule
\end{tabular}%
}
\end{table*}

\textbf{In-depth analysis of our PINN: } To study our PINN in detail, we vary different aspects of our method (e.g., the impact of individual loss/regularization terms, hidden layer configuration, batch size, and activation functions). At a time, we vary one focused aspect, and fix all other hyper-parameters as per the default setting prescribed above.

Firstly, we study the empirical convergence of the default setting of our method. Fig \ref{conv_ebvs} plots the convergence behaviour of each of the loss terms individually (supervised, EBV, and EBS). Our method, as shown, enjoys a good convergence. In Table \ref{ablation_hidden}, we report the performance of our method by varying the hidden layer configurations (e.g., $[100]$ denotes one hidden layer with 100 neurons, $[50, 100]$ denotes two hidden layers with 50, and 100 neurons respectively, and so forth). The maximum values for each row (corresponding to a metric) are shown in bold. We found that it suffices to use [50, 100, 200] configuration for a competitive performance.

In Table \ref{ablation_batchsz}, we vary the batch size in our method. We found a batch size of 64 to provide an optimal performance for our experiments. In Table \ref{ablation_ebv_ebs}, we study the effect of individual physics-based regularization terms used in our method. We found that using both instances of volume and surface zone based regularizers together leads to better performance as compared to either EBV or EBS in isolation.

In Table \ref{ablation_activationfn}, we vary the underlying activation functions throughout our model. While we observed the benefits of using ReLU, SiLU, and Mish over others, there is no clear winner. All three lead to competitive performance. But when it comes to consistent performance across batch sizes, we noticed from our experiments that ReLU is more robust. Thus, we could recommend using the basic ReLU as de facto in our experiments.

\textbf{Additional comparisons of our PINN against classical ML techniques: } In addition to DL, we also compare our method against the classical ML baselines (in an IID evaluation setting): i) Decision Tree (DT), ii) Random Forest (RF), and iii) Histogram Gradient Boosting (H-GBoost). When it comes to only the classical ML baselines, the performances are as per the expectation. For instance, with previous temperatures as inputs, the performance of DT, RF, and H-GBoost increases. However, being an ensemble learning method, RF performs superior to DT. At the same time, by virtue of boosting, among all the three classical methods, H-GBoost performs the best. We observe superior performance of our model against the classical baselines as well, as reported in Table \ref{iid_classical_ML}. H-GBoost, though competitive, is significantly slower for our studied case of multi-output regression.


\end{document}